\documentclass[conference]{IEEEtran}
\IEEEoverridecommandlockouts
\usepackage{cite}
\usepackage{amsmath,amssymb,amsfonts}
\usepackage[linesnumbered,ruled,vlined]{algorithm2e}
\usepackage{graphicx}
\usepackage{textcomp}
\usepackage{xcolor}
\usepackage{algpseudocode}
\usepackage{amsfonts}
\usepackage{amsthm}
\usepackage{stfloats}
\usepackage{enumerate}
\usepackage{cite}

\def\BibTeX{{\rm B\kern-.05em{\sc i\kern-.025em b}\kern-.08em
    T\kern-.1667em\lower.7ex\hbox{E}\kern-.125emX}}

\begin{document}
\setlength{\abovecaptionskip}{0pt}
\setlength{\belowcaptionskip}{0pt}
\title{Adaptive Federated Learning in Heterogeneous Wireless Networks with Independent Sampling}
\author{
	\IEEEauthorblockN{
		Jiaxiang Geng\IEEEauthorrefmark{1}, 
		Yanzhao Hou\IEEEauthorrefmark{1}\IEEEauthorrefmark{4}, 
		Xiaofeng Tao\IEEEauthorrefmark{1}\IEEEauthorrefmark{4}, 
		Juncheng Wang\IEEEauthorrefmark{2}, and 
		Bing Luo\IEEEauthorrefmark{4}} 
	\IEEEauthorblockA{\IEEEauthorrefmark{1} Beijing University of Posts and Telecommunications, Beijing, China}
	\IEEEauthorblockA{\IEEEauthorrefmark{2} Hong Kong Baptist University, Hong Kong, China}
    \IEEEauthorblockA{\IEEEauthorrefmark{4} Peng Cheng Laboratory, Shenzhen, China } 
    
    \thanks{This work was supported in part by the Joint Research Fund for Beijing Natural Science Foundation and Haidian Original Innovation under Grant L232001, in part by the National Key R\&D Program of China under Grant 2019YFE0114000, in part by the Major Key Project of PCL, in part by the 111 Project of China (No. B16006), and in part by the Fundamental Research Funds for the Central Universities under Grants 2242022k60006. Bing Luo is affiliated with Duke Kunshan University. This work was conducted while he was a Visiting Scholar at Peng Cheng Laboratory. Corresponding author: Bing Luo (bing.luo@dukekunshan.edu.cn)}
} 

\maketitle
\vspace*{-30pt}

\begin{abstract}
Federated Learning (FL) algorithms commonly sample a random subset of clients to address the straggler issue and improve communication efficiency. While recent works have proposed various client sampling methods, they have limitations in joint system and data heterogeneity design, which may not align with practical heterogeneous wireless networks. In this work, we advocate a new independent client sampling strategy to minimize the wall-clock training time of FL, while considering data heterogeneity and system heterogeneity in both communication and computation. We first derive a new convergence bound for non-convex loss functions with independent client sampling and then propose an adaptive bandwidth allocation scheme. Furthermore, we propose an efficient independent client sampling algorithm based on the upper bounds on the convergence rounds and the expected per-round training time, to minimize the wall-clock time of FL, while considering both the data and system heterogeneity. Experimental results under practical wireless network settings with real-world prototype demonstrate that the proposed independent sampling scheme substantially outperforms the current best sampling schemes under various training models and datasets.

\end{abstract}


\section{Introduction}
With rapid evolution of 5G, Internet of Things, and social networking applications, wireless network edge generates an exponential surge of data\cite{b1}. Centralized model training at wireless edge raises \emph{privacy concerns}\cite{b2}. As an appealing distributed machine learning (DML) scheme, Federated Learning (FL) allows multiple local clients to collaboratively train a global machine learning model without collecting their raw data. FL has broad applications in various fields, spanning computer vision, natural language processing, e-Healthcare, and smart home\cite{b3,b4,b5,10239352}.

Compared to traditional DML schemes, FL has two distinctive features: First, the data distribution is highly non-i.i.d. and unbalanced, known as \emph{data heterogeneity}, which can be detrimental to the convergence performance\cite{b20}. Second, the local clients typically have diverse communication and computation capabilities, known as \emph{system heterogeneity}, causing the straggler issue that slows down the training process~\cite{b21}. The impact of system heterogeneity on FL in wireless edge networks is more severe due to the limited communication bandwidth  resources shared by all the local clients\cite{b25}.

To address the network heterogeneity and improve communication efficiency, FL algorithms such as the standard FedAvg algorithm typically sample a \textit{subset} of clients and perform multiple local model updates\cite{b8}. Recent studies have provided theoretical convergence analysis for FL with \emph{client sampling}\footnote{Client sampling here means that all clients will participate in FL throughout the training rounds, with certain probability of participation in each round (not to casually sample a fixed subset of clients in each round), which guarantees the model convergence and unbiasedness.}\cite{b12,b22,b23}. However, these works have primarily focused on the data heterogeneity but neglecting the system heterogeneity, and therefore often exhibit slow convergence in terms of the wall-clock time. The reason is that these works may select a straggler under poor wireless channel condition, resulting in a long per-round time\cite{9488679}.
\begin{figure}[]
\vspace{-2em}
\centering
\includegraphics[width=0.48\textwidth]{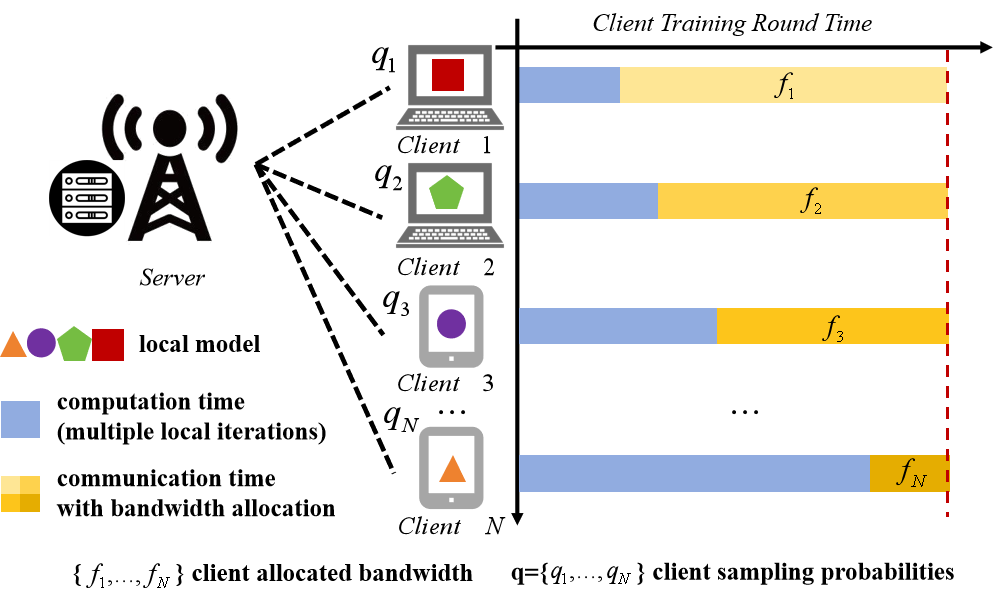}
\caption{A heterogeneous federated learning over wireless networks, where, clients with heterogeneous computation and communication capacities, as well as data distribution.}
\label{fig}
\end{figure}

In addressing both data and system heterogeneity, recent works \cite{b15,b,b26} have designed client sampling strategies in wireless networks to minimize FL convergence time. However, the convergence analysis in \cite{b15} is only valid for convex loss function. Furthermore, their sampling probabilities among clients are \emph{dependent}, and a \emph{fixed} number of participating clients are selected in each round. In practice, some FL clients may experience unexpected disconnections due to the fluctuation of wireless networks, while others may be unable to participate in FL training due to their higher-priority tasks, which make the participation of each client in FL training \emph{independent}. Although \cite{b} and \cite{b26} studied independent client sampling, they only considered the communication heterogeneity while neglecting the computation heterogeneity. As shown in Fig. 1, when FL clients perform multiple local iterations, computation heterogeneity also has a significant impact on the training round time, and should be jointly designed and optimized with the communication heterogeneity. Moreover, their convergence results rely on a uniform bounded gradient assumption, which may result in a loose bound, since the gradient norm usually decays over the training rounds. 

Motivated by the above discrepancies, in this work, we propose a new \emph{independent client sampling} strategy to minimize the FL convergence wall-clock time, while considering data heterogeneity and system heterogeneity in \textit{both} communication and computation. The main contributions of this paper are as follows:

\begin{itemize}
\item We derive a new convergence bound for FL with arbitrary and independent client sampling probabilities. Our convergence bound holds for general non-convex loss function without the stringent bounded gradient assumption as in\cite{b,b26}.
\item We propose an adaptive bandwidth allocation scheme for FL with independent client sampling. Our proposed approach jointly considers the impacts of heterogeneous computation and communication capacities, as well as the limited system bandwidth, to characterize the expected wall-clock time for each training round.
\item Building upon the derived convergence bound and the proposed adaptive bandwidth allocation scheme, we formulate an optimization problem on the independent sampling probabilities, to minimize the FL convergence wall-clock time with both data and system heterogeneity. We develop an efficient algorithm to approximately solve the non-convex optimization problem with low computational complexity.
\item We evaluate the performance of our proposed algorithm using a real-world hardware prototype under practical heterogeneous network settings.
Our experimental results under various learning models and datasets demonstrate that the proposed algorithm substantially reduces the FL wall-clock training time compared with the current best sampling schemes.
\end{itemize} 

\section{FL with Independent Client Sampling}
We first introduce a more general version of FedAvg with arbitrary and independent client sampling. Then, we derive a new convergence bound for non-convex loss functions under mild assumptions.
\subsection{Unbiased FL with Independent Client Sampling}
As depicted in Fig. 1, we consider a FL system at wireless edge, comprising $\mathcal{N}=1,...,N$ clients and one server. The objective of FL is for the clients to cooperate with each other with the assistance of the server, to find the optimal global machine learning model based on local data by solving:
\begin{equation}
\small
    \min_x F(x)=\sum_{n=1}^N a_n F_n(x) \label{flgoal}.
\end{equation}
Here, $F(x)$ represents the global loss function, and $F_n(x)$ represents the local loss function of client $n$. The aggregation weight $a_n=\frac{|D_n|}{|D|}$ is set as the data ratio of client $n$, where $|D_n|$ and $|D|$ represent the local and global data sizes, such that $\sum_{n=1}^N a_n=1$. 

We consider a more general version of FedAvg\cite{b8} with independent client sampling. We adopt the following adaptive aggregation rule in \cite{b15,b,b26} to ensure the aggregated global model remains unbiased:
\begin{equation}
\small
    x_{r+1} \gets x_r-\sum_{n=1}^{N}a_n\eta\frac{\mathbb{I}_r^n}{q_n}g_n(x_r),
\label{aggregation}
\end{equation}
where $x_r$ denotes the aggregated global model in round $r$. $x_r^n$ denotes the local model of client $n$ in round $r$. Each participating device leverages its local data to calculate its local stochastic gradient descent, denoted as $g_n(x_r)$, which may consist of multiple gradient computations with a learning rate $\eta >0$ on the global model $x_r$ obtained in the previous round. Each client $n$ is sampled by an independent Bernoulli trial $\mathbb{I}_r^n \sim \text{Bernoulli}(q_n)$ at each round $r$, where $0<q_n \le 1$ is the sampling probability of client $n$. Let $\mathbf{q} = \{q_1,\dots,q_N\}$. When client $n$ participates in round $r$, we have $\mathbb{I}_r^n=1$; otherwise, we have $\mathbb{I}_r^n=0$. Therefore, the expectation of the aggregated global model satisfies $E[x_{k+1}]=\sum_{n=1}^{N} a_n x_k^n$, which guarantees that the aggregated global model after independent client sampling is unbiased \cite{b15,b,b26}. In Algorithm~1, we summarize FL with independent client sampling. 

\begin{algorithm}[t]
\label{algorithm_1}
\small
 \caption{FL with Independent Client Sampling}
 \KwIn{Sampling probability $\!\textbf{q}\!=\!\{q_1,...,q_N\!\}$, initial model~$x_0$}
 \KwOut{Final model parameter $x_R$}
 \For{$r=0,1,2,..., R-1$}
 { 
At each client $n=1,2,...,N$, \textbf{do}: 
 
\quad Sample $\mathbb{I}_r^n \sim \text{Bernoulli}(q_n)$; \quad \quad // \emph{Sampling}

\quad Update gradients;  \quad \quad \quad \quad \quad// \emph{Computation}

\quad Send $-\frac{\eta\mathbb{I}_r^n}{q_n}g_n(x_r)$ to server;  \quad // \emph{Communication}

At the server, \textbf{do}:

\quad Update $\!x_{r+1} \!\gets\! x_r \!- \!\sum_{n=1}^{N}\! a_n \frac{\eta\mathbb{I}_r^n}{q_n}g_n(\!x_r\!)$; // \emph{Aggregation}

\quad Send $x_{r+1}$ to all clients; \quad \quad // \emph{Synchronization}
}
\end{algorithm}

\subsection{Convergence Analysis for Independent Client Sampling}
We present a new convergence bound for Algorithm 1 with non-convex loss functions. We make the following mild assumptions that are commonly adopted in existing FL works such as \cite{b24,b11,b27}.\\
\textbf{Assumption 1.} Lipschitz gradient: The difference between the local gradient $\nabla F_n(x)$ and $\nabla F_n(y)$ satisfies
\begin{center}
    $\left \|\nabla F_n\left(x\right)-\nabla F_n\left(y\right) \right \| \le L \left \|x-y \right \| , \forall x, y, \forall n\in\mathcal{N}.$
\end{center}
\textbf{Assumption 2.} Unbiased stochastic gradient with bounded variance: The stochastic local gradient $g_n(x)$ satisfies
\begin{center}
 $E\left[g_n\left(x\right)\right]=\nabla F_n\left(x\right), \forall{x},\forall n\in\mathcal{N}$
 \end{center}
 \begin{center}
 $E\left[\left \|g_n\left(x\right)-\nabla F_n\left(x\right) \right \|^2\right] \le \sigma^2. $
\end{center}
\textbf{Assumption 3.} Bounded gradient divergence: The difference between the local gradient $\nabla{F}_n(x)$ and the global gradient $\nabla{F}(x)$ satisfies
\begin{center}
$\left \|\nabla F_n\left(x\right)-\nabla F\left(x\right) \right \|^2 \le \varepsilon^2,\forall{x}, \forall{n}\in\mathcal{N}.$
\end{center}

The gradient divergence upper bound $\varepsilon^2$ captures the degree of data heterogeneity (i.e., non-i.i.d.) across clients. The following theorem provides a new convergence bound for Algorithm~1 with non-convex loss functions.

\textbf{Theorem 1.} Let $F^*:=\min_x F(x)$. If $N\sum_{n=1}^N \frac{a_n^2}{q_n} \le p$ and $\eta \le \frac{1}{4Lp}$, for a total of R training rounds, the global model sequence $\{x_r\}$ yielded by Algorithm 1 satisfies:
\begin{equation}
\small
\begin{aligned}
    \!\frac{1}{R}\!\sum_{r=0}^{R-1}\!E&[\left \| \nabla F(x_r) \right \|^2] \!\le \\
    &\! \frac{4(F(x_0)\!-\!F^*)}{\eta R}\!+\!4\eta L N(\varepsilon ^2\!+\!\sigma^2)\!\sum_{n=1}^N  \frac{a_n^2}{q_n}
\label{upperbound}
\end{aligned}
\end{equation}
\begin{IEEEproof}
We present a brief proof sketch here due to space constraints. The proof process follows the approach outlined in \cite{b24}, with the additional introduction of $a_n$ in our case. We first derive an upper bound for the difference between \(E[F(x_{r+1})]\) and \(E[F(x_r)]\). This is achieved by considering the local stochastic gradient descent (SGD) updates, the aggregation rule, and the $L$-smoothness of the global loss function $F$. Subsequently, we proceed to further upper bound the difference within the first step. We leverage the independence between client sampling and the randomness in data sampling of SGD, employing tools such as Jensen’s inequality and Young’s inequality. Finally, we sum the inequality derived in the previous two steps over round $r$ from 0 to \(R-1\), take the total expectation, and rearrange terms to obtain the convergence bound.
\end{IEEEproof}

From the results in Theorem 1, we can draw the following two theoretical insights. Note that our convergence bound in (3) holds for arbitrary and independent client sampling probabilities $\mathbf{q}$, i.e., $\sum_{n=1}^N q_n \in (0,N]$.

We show that the upper bound on the averaged expected global gradient decreases as the independent sampling probabilities increase. In other words, when all devices have participation probabilities $q_n=1$, i.e., full participation, the upper bound reaches its minimum value, and the required total number of rounds $R$ to reach a preset convergence threshold $\xi$ is minimized.  The upper bound in \eqref{upperbound} further implies that in order to obtain an unbiased global model, all clients need to participate with a positive probability for model convergence, i.e., $q_n \neq 0$, for all $n$. This is because when $q_n \to 0$, it will take infinite number of rounds $R$ to ahieve convergence.



\textbf{Remark 1}: Although a larger value of \(q_n\) results in a faster convergence rate (e.g., a smaller number of rounds \(R\) to reach the same convergence threshold \(\xi\)), it does not necessarily mean a reduction in the total wall-clock convergence time. In a round where more devices participate, each participating device may be allocated with less bandwidth, leading to longer communication time. Moreover, the involvement of straggler devices, due to limited computational capabilities, can further increase the round time. These motivate us to answer the following question: \textit{How to optimize the independent client sampling probabilities to minimize the wall-clock time for the global model to reach a preset convergence threshold?}

\section{Problem Formulation}
In this section, we first introduce the proposed adaptive bandwidth allocation scheme for FL with independent client sampling, while considering the data heterogeneity and the system heterogeneity in both communication and computation. Then, we present our formulated optimization problem on the independent sampling probabilities to minimize the wall-clock training time, subject to a convergence threshold constraint.
\subsection{Adaptive Bandwidth Allocation} 
As shown in Fig. 1, we consider both heterogeneous communication and computation time. For the communication time, due to the system bandwidth limitation and wireless interference, we assume that the selected clients are allocated with orthogonal bandwidths. For the computation time, we assume it is a constant and is measurable for each client. 

Let $\tau_n$ be the computation time of client $n$, $t_n$ be its communication time with one unit bandwidth, and $\hat{f}_i^{(r)}$ be its allocated bandwidth at round $r$. We can show that for any independent sampling probability \textbf{q}, a minimum round time $\hat{T}^{(r)}(\textbf{q})$ is obtained when the sampled clients complete their training round $r$ at the same time\cite{b15}. The minimum round-time $\hat{T}^{(r)}(\textbf{q})$ can be expressed as:
\begin{equation}
\small
    \hat{T}^{(r)}(\textbf{q})=\tau_n+\frac{t_n}{\hat{f}^{(r)}_n},\forall n \in \mathcal{K}^{(r)}(\textbf{q}),     \label{design}
\end{equation}
where $\mathcal{K}^{(r)}(\mathbf{q})$ represents the set of devices participating in round $r$ under sampling probability \textbf{q}. Note that size and elements of $\mathcal{K}^{(r)}(\mathbf{q})$ vary across rounds, making the allocation of $\hat{f}_n^{(r)}$ to minimize the total wall-clock training time a highly intricate problem. For analytical convenience, we consider the expected bandwidth $f_n^{(r)}$ allocation instead and approximate $\hat{T}^{(r)}(\textbf{q})$ with $T^{(r)}(\textbf{q})$ under the expected bandwidth. For a given round $r$, from \eqref{design}, the expected total bandwidth $f_{tot}$ can be expressed~as:
\begin{equation}
\small
    f_{tot}=\sum_{n=1}^N q_n f_n^{(r)}=\sum_{n=1}^N q_n \frac{t_n}{T^{(r)}(\textbf{q})-\tau_n}.
    \label{ftot}
\end{equation}
The total training time $T_{\text{tot}}(\mathbf{q}, R)$ after $R$ rounds can therefore be expressed as:
\begin{equation}
\small
    T_{tot}(\textbf{q},R)=\sum_{r=1}^R T^{(r)}(\textbf{q}).
    \label{Ttot}
\end{equation}
\subsection{Problem Formulation}
Our goal is to optimize the independent client sampling probabilities \textbf{q}, to minimize the expected total training time $E[T_{\text{tot}}(\mathbf{q}, R)]$, while ensuring that the average squared norm of global gradient $\frac{1}{R}\sum_{r=0}^{R-1}E[||\nabla F(x_R)||^2]$ is under a preset convergence threshold $\xi$ after $R$ rounds. This translates into the following optimization problem:
\begin{align}
\small
  \textbf{P1:} \quad &\min_{\textbf{q},R} E[T_{tot}(\textbf{q},R)] \nonumber\\
  \text{s.t.} \quad & \frac{1}{R}\sum_{r=0}^{R-1}E[\left \| \nabla F(x_r) \right \|^2] \le \xi, \quad R\in Z^+, \label{P1}\\
  &0<q_n \le 1, \quad \forall n \in \mathcal{N}. \nonumber
\end{align}

$\textbf{P1}$ is a joint and non-convex optimization problem and is difficult to solve. First, the global loss function $F(x)$ is generally defined and it is typically impossible to predict how $\mathbf{q}$ and $R$ affect the global loss before actually training the model. Second, the analytical expression of the round time $T^{(r)}(\textbf{q})$ is not available.
\addtolength{\topmargin}{.02in}
\section{Optimization for Independent Client Sampling}
In this section, we develop an efficient algorithm to solve \textbf{P1}. We first obtain an upper bound on the expected round time $E[T^{(r)}(\textbf{q})]$. Then, we formulate an approximate problem of \textbf{P1} based on the convergence upper bound in Theorem 1. Finally, we propose an algorithm to solve the approximate problem with low computational complexity. 

\subsection{Bounding the Expected Round Time $E[T^{(r)}(\textbf{q})]$}
From \eqref{Ttot}, if we can obtain the expected round time $E[T^{(r)}(\textbf{q})]$, multiplying it by the total number of rounds $R$ will give us an approximation of the total training time, in the objective of \textbf{P1}. Without loss of generality, we assume the $N$ clients are ordered such that their computation time satisfies $\tau_1 \le \tau_2 \le ... \le \tau_N$. In the following theorem, we provide an upper bound for $E[T^{(r)}(\textbf{q})]$.

\textbf{Theorem 2.} The expected round time $E[T^{(r)}(\textbf{q})]$ is upper bounded by
\begin{equation}
\small
E[T^{(r)}(\textbf{q})] \le \frac{\sum_{n=1}^N q_n t_n}{f_{tot}}+E[\max{\tau_n}],
\label{theorem2}
\end{equation}
where $E[\max{\tau_n}]$ is the expected maximum computation time among the clients, given by:
\begin{equation}
\small
E[\max{\tau_n}]=q_N \tau_N + \sum_{n=1}^{N-1} \prod_{i=n+1}^{N} (1-q_i)q_n \tau_n \le \sum_{n=1}^N q_n \tau_n.
\label{maxtau}
\end{equation}
\begin{IEEEproof}
It can be shown that the probability of client $n$ being the slowest one is $\prod_{i=n+1}^{N} (1-q_i)q_n$. This is because client $n$ is the slowest implying that clients $n+1$ to $N$ did not participate in the training. Due to $\prod_{i=n+1}^{N} (1-q_i) \le 1$, the upper bound of $E[\max{\tau_n}]$ can be established. Furthermore, we introduce $\max{\tau_n}$ to find the maximum value of (\ref{ftot}):
\begin{equation}
\small
f_{tot}=\sum_{n=1}^N q_n \frac{t_n}{T^{(r)}(\textbf{q})-\tau_n} \le \frac{\sum_{n=1}^N q_nt_n}{T^{(r)}(\textbf{q})-\max{\tau_n}}.
\end{equation}
We move $f_{tot}$ to the right side of the equation and $T^{(r)}(\textbf{q})$ to the left side, then take the expectation on both sides, resulting in \eqref{theorem2}.
\end{IEEEproof}
From the results in Theorem 2, we can derive an upper bound on $E[T^{(r)}(\textbf{q})]$ as follows:
\begin{equation}
\small
E[T^{(r)}(\textbf{q})] \le \sum_{n=1}^N q_n(\frac{t_n}{f_{tot}}+\tau_n).
\end{equation}

\subsection{Approximate Optimization Problem for \textbf{P1}}
As a result, we can approximate the optimal solution to \textbf{P1} by minimizing an upper bound of the problem with Theorem 1 and 2, which can be expressed as follows:
\begin{small}
\begin{align}
\small
  \textbf{P2:} \quad &\min_{\textbf{q}} \frac{\alpha}{\beta-\sum_{n=1}^N\frac{a_n^2}{q_n}} \sum_{n=1}^N q_n(\frac{t_n}{f_{tot}}+\tau_n) \nonumber \\
  \text{s.t.} \quad &\frac{a_n^2N}{\beta} < q_n \le 1, \quad \forall n \in \mathcal{N} \label{P2}
\end{align}
\end{small}
where $\alpha=\frac{F(x_0)-F^*}{\eta^2LN(\varepsilon ^2+\sigma^2)}>0$ and $\beta=\frac{\xi}{4\eta LN(\varepsilon ^2+\sigma^2)}>0$ are constants. Here we additionally require two conditions: $\xi > \varepsilon ^2+\sigma^2$ and $\eta \le \min\{\frac{1}{4LQ},\frac{1}{4LN^2}\}$ to ensure that the denominator of \textbf{P2} is greater than zero, which can be derived from the constraints of Theorem 1.

\textbf{Remark 2}: Here, we are imposing a tighter lower bound constraint \eqref{P2} on the samping probabilities $q_n$ for each device, which controls that if a device possesses a larger amount of data, this lower bound for participation probability needs to be higher. An important special case is when the data is evenly distributed among all devices, i.e., $a_n=\frac{1}{N}$. In this case, can obtain $\frac{4\eta L (\varepsilon ^2 + \sigma^2)}{\xi} < q_n \le 1$, indicating that the lower bound of the sampling probability for this device is related to both the degree of non-iidness $\varepsilon ^2$ and gradient computation precision~$\sigma^2$.

\subsection{Optimization Algorithm for \textbf{q}}
Solving \textbf{P2} poses two major challenges. One involves the presence of unknown parameters, $\alpha$ and $\beta$. The other challenge stems from the non-convex nature of \textbf{P2}.

We first propose a method to estimate the unknown parameters $\alpha$ and $\beta$. This approach leverages the convergence upper bound derived in (\ref{upperbound}), together with implementing Algorithm 1 using two reference sampling schemes: uniform sampling \(\textbf{q}_1\) with \(q_n = \frac{1}{N}\) and full sampling \(\textbf{q}_2\) with \(q_n = 1\).

We only run a limited number of rounds to reach a predefined estimation loss \(F_s\) instead the convergence threshold \(\xi\). Let's assume that \(R_1\) rounds are required to reach the loss \(F_s\) with \(\textbf{q}_1\), and \(R_2\) rounds are needed to achieve the same loss \(F_s\) with \(\textbf{q}_2\). Additionally, denote the sum \(\sum_{n=1}^N \frac{a_n^2}{q_n}\) as \(C_1\) under \(\textbf{q}_1\), and \(C_2\) under \(\textbf{q}_2\). We can derive the following relationships:
\begin{equation}
\small
  \begin{aligned}
  R_1=\frac{\alpha}{\beta-C_1},  \ \ \ \ R_2=\frac{\alpha}{\beta-C_2}.
  \end{aligned}
\end{equation}
Therefore, we can get:
\begin{equation}
\small
  \beta=\frac{R_1C_1-R_2C_2}{R1-R2}, \ \ \ \ \alpha=R_1R_2\frac{C_1-C_2}{R_1-R_2}.
\label{solvedalphabeta}
\end{equation}

The overall estimation process corresponds to lines 1–4 of Algorithm 2.

Next, we convert \textbf{P2} to a convex problem. We decompose the objective of \textbf{P2} into two multipliers: $\alpha/({\beta-\sum_{n=1}^N\frac{a_n^2}{q_n}})$ and $\sum_{n=1}^N q_n(\frac{t_n}{f_{tot}}+\tau_n)$. We treat the latter term as a control variable $M$, and use a convex function to approximate the former term. Define the control variable $M$ as:
\begin{equation}
\small
M:=\sum_{n=1}^N q_n(\frac{t_n}{f_{tot}}+\tau_n).
\label{M}
\end{equation}
where $M_{min}=N \cdot \min_{n \in \mathcal{N}}( \frac{a_n^2 N}{\beta} (\frac{t_n}{f_{tot}}+\tau_n)) \le M \le M_{max} = N \cdot \max_{n \in \mathcal{N}}( \frac{t_n}{f_{tot}}+\tau_n)$. The first term in \textbf{P2} can be convert to a convex function using the property that the harmonic mean is no greater than the arithmetic mean, given~by:
\begin{small}
\begin{align}
\frac{\alpha}{\beta-\sum_{n=1}^N \frac{a_n^2}{q_n}} &\!=\! \frac{N\alpha}{(\beta-\frac{a_1^2N}{q_1})+(\beta-\frac{a_2^2N}{q_2})+...+(\beta-\frac{a_N^2N}{q_N})} \nonumber \\
&\!\le\! \sum_{n=1}^N\frac{\alpha q_n}{N\beta q_n-a_n^2N^2} \label{inequality}
\end{align}
\end{small}
Note that when $\frac{a_n^2N}{\beta} < q_n \le 1, \forall n \in \mathcal{N}$, the right-hand side of (\ref{inequality}) is a convex function. The equality holds if and only if $\frac{\alpha q_i}{\beta q_i-a_i^2N}=\frac{\alpha q_j}{\beta q_j-a_j^2N}, \forall i,j \in \mathcal{N}, i \neq j$.

Using \eqref{M} and \eqref{inequality}, we convert \textbf{P2} to the following optimization problem:
\begin{small}
\begin{align}
  \textbf{P3:} \quad &\min_{\textbf{q}}\sum_{n=1}^N\frac{\alpha q_n}{N\beta q_n-a_n^2N^2} M \nonumber\\
  \text{s.t.} \quad & M=\sum_{n=1}^N q_n(\frac{t_n}{f_{tot}}+\tau_n), \label{P3}\\
  &\frac{a_n^2N}{\beta} < q_n \le 1, \quad \forall n \in \mathcal{N} \nonumber
\end{align}
\end{small}
For any fixed feasible value of $M \in [M_{min},M_{max}]$, \textbf{P3} is convex in \textbf{q}. Thus, we can solve \textbf{P3} with a convex optimization tool, such as CVX, while tackling the problem by employing a linear search approach with a constant step size of $\epsilon$ over the range $[M_{min}, M_{max}]$\cite{b15}. This optimization process corresponds to lines 5-8 of the proposed Algorithm~2.
\begin{algorithm}[t]
\label{algorithm_2}
\small
 \caption{Approximately Optimal Independent Sampling for FL with Data and System Heterogeneity}
\KwIn{$N,\tau_n,t_n,f_{tot},a_n=|D_n|/|D|,x_0,F_s,\epsilon$}
 \KwOut{Approximation of $\textbf{q}^*$}

Server runs Algorithm 1 with uniform sampling $\textbf{q}_1$ and full sampling $\textbf{q}_2$, respectively;

 Server recored $R_1$ and $R_2$, when reaching $F_s$;

Calculate $\alpha$ and $\beta$ using (\ref{solvedalphabeta});

\For{$M(\epsilon)=M_{min},M_{min}+\epsilon,M_{min}+2\epsilon,...,M_{max}$}{
Substitute $N,\tau_n,t_n,f_{tot},a_n,M(\epsilon),\alpha,\beta$ into \textbf{P3};

Solve \textbf{P3} via CVX, and obtain $\textbf{q}^*(M(\epsilon))$;
}
\textbf{return} $\textbf{q}^*(M^*(\epsilon))=arg min_{M(\epsilon)} \textbf{q}^*(M(\epsilon))$
\end{algorithm}
\section{Experiment and Results}
\subsection{Experiment Setup on Real-world Mobile Devices}
Our experimental setup comprises a model aggregation server and several types of real-world mobile devices. On the FL server side, we employ an NVIDIA RTX 3090. On the FL client side, we utilize five types of devices: Xiaomi 12S, Honor Play 6T, Honor 70 smartphones, a MacBook Pro 2018 laptop, and a Teclast M40 tablet. We develope a tool for model training on Android smartphones and laptops. We mirror the parameters of the hardware to virtual devices in the simulated environment. To be specific, we conduct experiments with a total of 100 mobile endpoints mirrors, i.e., $N=100$, with 20 devices for each of the aforementioned mobile device types. The communication between FL clients and the server is established with an assumed total server bandwidth. The connection from mobile endpoints to the server is made through an indoor Wi-Fi 5 base station.

\subsection{FL Tasks and Parameter Settings}
We conduct three FL tasks, including~two~image classification tasks: CNN@MNIST and MobileNet@CIFAR10, and a next-word prediction task: LSTM@Shakespeare. We use a stochastic gradient descent (SGD) batch size of 32, with each client performing 10 local iterations. For the two image classification tasks, the learning rate is set to 0.01, while for the next-word prediction task, the learning rate is set to 0.8. The data on each device is non-i.i.d., following a Dirichlet distribution with a concentration parameter of 0.8\cite{b18}. Unless otherwise specified, the total bandwidth is set at 100Mbps.

\subsection{Results}
We compare the performance of our proposed client sampling scheme with four benchmarks: full sampling with $q_n=1$, fixed sampling with a predefined fixed value of $q_n=0.2$, uniform sampling with $q_n=1/N$, and weighted sampling with $q_n=|D_n|/|D|$. In each of these client sampling methods, all devices participate independently. After conducting multiple experiments and averaging the results, we present the experimental findings in Fig. 2 to 4 and Table 1. In our analysis, we focus on wall-clock time, and all time measurements are recorded in hours.

From the experimental results, it is evident that in the CNN@MNIST task, the proposed sampling scheme achieves the shortest time to reach the target accuracy. In comparison, full sampling takes 7.35 times longer due to bandwidth constraints. Weighted sampling outperforms uniform sampling, but they still require 1.57 and 1.68 times more time, respectively, than our proposed scheme. For larger model tasks like MobileNet@CIFAR10, where more client devices are needed to participate, the fixed sampling scheme with $q_n=0.2$ outperforms uniform and weighted sampling. Our proposed scheme automatically identifies this pattern based on the values of $\alpha$ and $\beta$. Similarly, in the next-word prediction task, LSTM@Shakespeare, the proposed scheme also obtain results that outperform all the benchmarks.
\begin{table*}[]
\centering
\caption{Training hours for three FL tasks of proposed sampling scheme with four benchmarks}
\begin{tabular}{c|c|c|c|c|c|c}
\hline
FL Task           & Target Accuracy & Proposed Sampling & Full Sampling     & Fixed Sampling    & Weighted Sampling & Uniform Sampling  \\ \hline
CNN@MNIST         & 95.00\%         & \textbf{2.04h}    & 14.99h ($\times$7.35)  & 7.31h ($\times$3.58)   & 3.21h ($\times$1.57)   & 3.42h ($\times$1.68)   \\ \hline
MobileNet@CIFAR10 & 60.00\%         & \textbf{115.14h}  & 362.03h ($\times$3.14) & 164.35h ($\times$1.43) & 230.76h ($\times$2.00) & 241.24h ($\times$2.10) \\ \hline
LSTM@Shakespeare  & 48.00\%         & \textbf{11.68h}   & 322.87h ($\times$27.64) & 130.86h ($\times$11.20) & 23.89h ($\times$2.05)  & 28.24h ($\times$2.42)  \\ \hline
\end{tabular}
\end{table*}

\begin{figure*}
\vspace{-1em}
\begin{minipage}[t]{0.245\linewidth}
\centering
\includegraphics[width=1\textwidth]{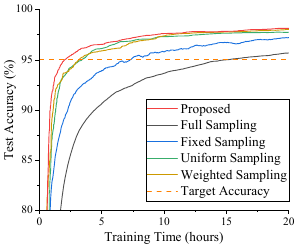}
\caption{CNN@MNIST}
\label{cnn}
\end{minipage}
\begin{minipage}[t]{0.245\linewidth}
\centering
\includegraphics[width=1\textwidth]{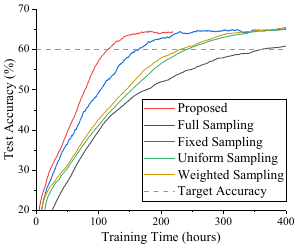}
\caption{MobileNet@CIFAR10}
\label{mobile}
\end{minipage}
\begin{minipage}[t]{0.245\linewidth}
\centering
\includegraphics[width=1\textwidth]{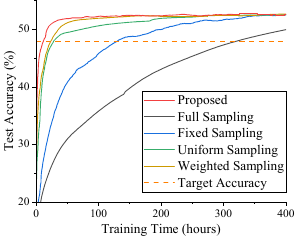}
\caption{LSTM@Shakespeare}
\label{lstm}
\end{minipage}
\begin{minipage}[t]{0.245\linewidth}
\centering
\includegraphics[width=1\textwidth]{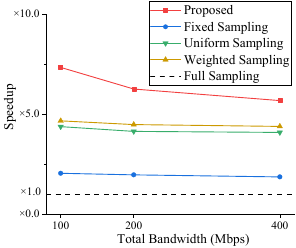}
\caption{Speedup with different $f_{tot}$}
\label{lstm}
\end{minipage}
\vspace{-0.5em}
\end{figure*}
Furthermore, we trained a CNN on an MNIST dataset to validate the proposed sampling under different total bandwidths \(f_{tot}\). As shown in Fig. 5, compared with full sampling, the proposed sampling can speed up convergence by \(\times 7.35\), \(\times 6.26\), and \(\times 5.68\) with 100Mbps, 200Mbps, and 400Mbps. It can be observed that the proposed method outperforms all the benchmarks, especially showing further improvements in challenging communication conditions.

\section{Conclusion}
In this paper, we consider minimizing the wall-clock training time of FL with independent client sampling in heterogeneous wireless networks. We start by conducting convergence analysis under arbitrary and independent device participation over non-i.i.d. data. We propose an adaptive bandwidth allocation strategy to minimize the per-round training time, considering the system heterogeneity in both communication and computation. Based on the upper bounds on the convergence rounds and the per-round training time, we propose an efficient independent client sampling scheme to minimize the wall-clock training time subject to a constraint on a preset convergence threshold. Extensive experimental results using a real-world prototype over various learning models and datasets demonstrate the effectiveness of the proposed sampling scheme over the current best alternatives.

\bibliographystyle{unsrt}
\bibliography{ref}

\end{document}